\documentclass[11pt]{article}

\usepackage[margin=1in]{geometry}
\usepackage{amsmath, amssymb}
\usepackage{graphicx}
\usepackage{booktabs}
\usepackage{hyperref}

\title{Deep Learning--Accelerated Surrogate Optimization for High-Dimensional Well Control in Stress-Sensitive Reservoirs}

\author{
Mahammad Valiyev$^{1}$, 
Jodel Cornelio$^{1}$, 
Behnam Jafarpour$^{1}$ \\
\vspace{0.2cm}
$^{1}$University of Southern California
}

\date{}

\begin{document}

\maketitle

\begin{abstract}
Production optimization in stress-sensitive unconventional reservoirs is governed by a nonlinear trade-off between pressure-driven flow and stress-induced degradation of fracture conductivity and matrix permeability. While higher drawdown enhances short-term production through stronger pressure gradients, it accelerates permeability loss via increased effective stress, leading to reduced long-term recovery. Identifying optimal, time-varying control strategies in this setting requires repeated evaluations of fully coupled flow–geomechanics simulators, making conventional optimization computationally prohibitive.

In this work, we propose a deep learning–based surrogate optimization framework for high-dimensional well control in stress-sensitive reservoirs. Unlike prior surrogate-based approaches that rely on predefined control parameterizations or generic sampling strategies, the proposed method treats well control as a fully continuous, high-dimensional optimization problem and introduces a problem-informed sampling strategy that aligns the training data distribution with trajectories encountered during optimization. A neural network proxy is trained to approximate the mapping between time-varying bottomhole pressure (BHP) trajectories and cumulative production using data generated from a field-scale coupled flow–geomechanics model.

The trained proxy is embedded within a constrained gradient-based optimization workflow, enabling rapid evaluation of candidate control strategies. Across multiple initialization scenarios, proxy-based optimization achieves agreement with full-physics solutions within a few percent in cumulative production, while reducing computational cost by up to three orders of magnitude. Analysis of failure cases shows that discrepancies are primarily associated with trajectories near the boundary of the training distribution and with local optimization effects, highlighting the importance of distribution-aware sampling.

The proposed framework demonstrates that surrogate models can be made both scalable and reliable for high-dimensional, PDE-constrained optimization by explicitly coupling data generation with the structure of the optimization problem. While demonstrated on a coupled flow–geomechanics system, the methodology is broadly applicable to other computationally intensive control problems in scientific and engineering domains.

\end{abstract}

\section{Introduction}

Unconventional oil and gas reservoirs have become a major component of global energy supply, enabled by advances in horizontal drilling and multi-stage hydraulic fracturing (Morrill and Miskimins, 2012; Patzek et al., 2013). Despite their economic significance, production from these systems is typically characterized by rapid decline following an initial high-rate period, resulting in reduced recovery efficiency and shortened productive life (Mirani et al., 2018; Kumar et al., 2018). A key mechanism underlying this behavior is the stress sensitivity of reservoir media, whereby depletion-induced increases in effective stress lead to degradation of fracture conductivity and matrix permeability (Liu et al., 2013; Wilson and Hanna Alla, 2017). Consequently, production performance in unconventional reservoirs is governed by tightly coupled interactions between fluid flow and geomechanical effects (Settari et al., 2002; Sun et al., 2021).

In stress-sensitive systems, well performance is strongly influenced by pressure drawdown, which controls both the pressure gradient driving fluid flow and the magnitude of stress-induced permeability loss. Increasing drawdown enhances production rates by increasing the pressure gradient between the reservoir and the wellbore; however, it simultaneously accelerates permeability degradation through increased effective stress (Sun et al., 2021; Zhao et al., 2022). Conversely, conservative drawdown strategies preserve permeability but limit production rates. These competing mechanisms imply the existence of an optimal, time-varying drawdown policy that balances short-term production gains against long-term reservoir integrity (Quintero and Devegowda, 2015; Rojas and Lerza, 2018). Identifying such policies requires joint consideration of nonlinear flow dynamics and geomechanical feedback, as well as systematic optimization over high-dimensional control spaces.

A substantial body of prior work has investigated drawdown management using both field data analysis and numerical simulation. Early studies typically relied on simplified or flow-only models, neglecting geomechanical effects or representing them in a limited manner (Almasoodi et al., 2020; Wilson and Hanna Alla, 2017). More recent work has incorporated coupled flow--geomechanics formulations to better capture stress-dependent permeability evolution and fracture closure behavior (Kumar et al., 2018; Zhao et al., 2022). While these studies provide valuable insights into the trade-offs between aggressive and conservative drawdown strategies, they are generally limited to evaluating a small number of predefined control trajectories, often assuming monotonic or linear pressure decline (Quintero and Devegowda, 2015; Almasoodi et al., 2020). Such approaches significantly restrict the search space and do not guarantee optimality in complex, nonlinear systems.

A more rigorous formulation treats well control as a continuous, time-dependent optimization problem, in which bottomhole pressure or flow rate is adjusted dynamically to maximize cumulative production or economic objectives (Yu and Jafarpour, 2022). However, solving this problem using high-fidelity reservoir models remains computationally prohibitive. In unconventional reservoirs, each evaluation requires a fully coupled flow--geomechanics simulation, which is computationally expensive due to strong nonlinear coupling between fluid flow and rock deformation (Settari et al., 2002; Tran et al., 2009). As a result, direct gradient-based optimization may require hundreds of simulator evaluations, leading to total runtimes on the order of hours to days for a single optimization instance (Valiyev et al., 2025). This computational bottleneck limits the practical applicability of rigorous optimization methods in field-scale settings.

To address this challenge, recent research has explored the use of reduced-order models and data-driven surrogate models to approximate the response of high-fidelity simulators (Yu and Jafarpour, 2022; Almasoodi et al., 2020). These approaches aim to replace expensive simulator evaluations with computationally efficient approximations, enabling faster optimization and uncertainty analysis. Among these methods, deep learning models have shown strong capability in approximating high-dimensional, nonlinear mappings. However, their application to well control optimization in stress-sensitive unconventional reservoirs remains limited, particularly in settings involving fully coupled flow--geomechanics physics and high-dimensional, time-varying control strategies.

In our previous work (Valiyev et al., 2025), we developed a field-scale framework for optimizing drawdown strategies using a fully coupled flow--geomechanics model. That study demonstrated that optimal, non-monotonic control trajectories can significantly outperform conventional predefined strategies, highlighting the importance of balancing early-time production with long-term permeability preservation. However, despite its physical rigor, the approach required extensive computational effort due to repeated evaluations of a high-fidelity simulator within the optimization loop, limiting its scalability.

In this work, we build upon that foundation and propose a deep learning–accelerated surrogate optimization framework for high-dimensional well control in stress-sensitive unconventional reservoirs. The key idea is to replace expensive simulator evaluations within the optimization loop with a neural network proxy trained on data generated from a coupled flow–geomechanics model. Unlike conventional approaches that rely on predefined control parameterizations, the proposed formulation treats drawdown as a continuous, high-dimensional control problem, enabling more flexible and expressive optimization.

From a machine learning perspective, this work addresses a central challenge in surrogate-based optimization under distribution shift: ensuring that the learned proxy remains accurate in regions of the input space explored by the optimizer. To address this, we introduce a problem-informed sampling strategy that aligns the training data distribution with trajectories encountered during optimization. Rather than relying on purely random or uniform sampling, the proposed approach generates physically realistic and optimization-relevant control trajectories, reducing extrapolation error and improving data efficiency.

The trained proxy model is embedded within a constrained, gradient-based optimization framework, enabling rapid evaluation of candidate control strategies and efficient navigation of the feasible control space. To ensure reliability, proxy-derived solutions are validated against the full-physics simulator, and discrepancies are analyzed to identify failure modes associated with distribution mismatch and local optimization effects.

The main contributions of this work are as follows:
\begin{enumerate}
\item A formulation of well control as a continuous, high-dimensional optimization problem for stress-sensitive reservoirs, moving beyond predefined drawdown strategies.
\item A problem-informed sampling strategy that aligns training data with the optimization search space, improving proxy robustness and reducing extrapolation-driven error.
\item A deep learning surrogate model for coupled flow--geomechanics systems, enabling efficient approximation of complex, nonlinear reservoir behavior.
\item A scalable optimization framework that achieves substantial computational speedup while maintaining agreement with high-fidelity simulator results.
\end{enumerate}

While the methodology is demonstrated for a coupled flow--geomechanics system, the proposed framework is broadly applicable to high-dimensional, PDE-constrained optimization problems in scientific and engineering domains where repeated simulator evaluations are computationally limiting.

\section{Methodology}

\subsection{Problem Formulation}

In stress-sensitive unconventional reservoirs, production performance is governed by the interaction between pressure-driven flow and stress-dependent permeability degradation. The objective of this study is to determine an optimal, time-varying bottomhole pressure (BHP) control policy that maximizes cumulative oil production over a given production horizon while satisfying operational constraints.

We consider a discretized control formulation in which the decision variable is a sequence of bottomhole pressure values defined over $T$ control intervals:
\[
\mathbf{u} = [p_1, p_2, \dots, p_T] \in \mathbb{R}^T,
\]
where $p_t$ denotes the bottomhole pressure at control step $t$, and $T$ is the total number of control steps. The discretization is non-uniform in time, with higher resolution during early production to capture transient effects and coarser resolution at later times.

The objective is to maximize cumulative oil production:
\[
\max_{\mathbf{u}} \; J(\mathbf{u}) = Q_{\mathrm{oil}}^{\mathrm{cum}}(\mathbf{u}),
\]
where $Q_{\mathrm{oil}}^{\mathrm{cum}}(\mathbf{u})$ is obtained from a fully coupled flow--geomechanics simulator.

The optimization is subject to operational constraints.

\paragraph{(1) Bounds on bottomhole pressure}
\[
p_{\min} \le p_t \le p_{\max}, \qquad \forall t = 1,\dots,T,
\]
where $p_{\min}$ and $p_{\max}$ represent minimum and maximum allowable BHP values.

\paragraph{(2) Step-change constraints}
\[
|p_t - p_{t-1}| \le \Delta p_{\max}, \qquad \forall t = 2,\dots,T,
\]
where $\Delta p_{\max}$ limits abrupt changes in control actions to ensure operational feasibility.

\paragraph{(3) Optional monotonicity constraint}
\[
p_t \le p_{t-1}, \qquad \forall t = 2,\dots,T,
\]
which enforces non-increasing drawdown trajectories. This constraint is applied selectively to analyze the effect of restricted control flexibility.

The mapping $J(\mathbf{u})$ is not available in closed form and is defined implicitly through a nonlinear, high-fidelity coupled flow--geomechanics simulator. Each function evaluation requires a full simulation, making direct optimization computationally expensive.

\subsection{Surrogate Reformulation}

To alleviate this computational burden, the optimization problem is reformulated using a data-driven surrogate model. Specifically, we approximate the simulator response using a neural network:
\[
J(\mathbf{u}) \approx \hat{J}(\mathbf{u}) = \hat{F}(\mathbf{u}),
\]
where $\hat{F}$ is a trained neural network mapping control trajectories to cumulative oil production.

The surrogate-based optimization problem becomes
\[
\max_{\mathbf{u}} \; \hat{J}(\mathbf{u})
\]
subject to the same operational constraints.

This reformulation enables efficient evaluation of candidate control trajectories and allows the use of gradient-based optimization methods with significantly reduced computational cost.

\subsection{High-Dimensional Control Perspective}

The control vector $\mathbf{u} \in \mathbb{R}^T$ defines a high-dimensional optimization problem, where $T$ corresponds to the number of control steps (typically $T \approx 20$ in this study). Although moderate in size, the control space exhibits strong nonlinear interactions due to the coupled physics of flow and geomechanics.

Unlike conventional approaches that restrict optimization to a small set of predefined parameterizations (e.g., linear or monotonic decline), the present formulation treats well control as a continuous, high-dimensional optimization problem. This enables the identification of non-trivial control trajectories that adapt dynamically to reservoir behavior.

However, this increased flexibility also introduces challenges for surrogate modeling. In particular, the accuracy of the proxy model depends critically on the distribution of training data relative to the regions explored during optimization. This motivates the development of problem-informed sampling strategies, described in subsequent sections.

\subsection{Coupled Flow--Geomechanics Model}

A field-scale, fully coupled flow--geomechanics model is employed to capture the interaction between pressure depletion, stress evolution, and permeability degradation in an unconventional reservoir containing a multi-stage hydraulically fractured horizontal well.

In stress-sensitive formations, production-induced pressure depletion increases effective stress, leading to a reduction in both matrix permeability and fracture conductivity. These changes directly impact fluid flow and, consequently, production performance. Accurate representation of this feedback requires a coupled formulation in which flow and geomechanical responses are solved consistently.

\subsubsection{Coupling Strategy}

The simulator adopts a sequential iterative coupling scheme. At each time step:
\begin{enumerate}
    \item The flow model is solved to update pressure and saturation fields.
    \item The resulting pressure field is passed to the geomechanical model to compute stress and deformation.
    \item Updated stresses are used to modify permeability and fracture conductivity.
    \item The updated properties are returned to the flow model for the next iteration.
\end{enumerate}

This procedure captures the two-way coupling between fluid flow and mechanical deformation while maintaining computational tractability for field-scale simulations.

\subsubsection{Stress-Dependent Permeability}

Permeability evolution is modeled using an exponential relationship with effective stress:
\[
k = k_0 \exp(-\alpha \Delta \sigma_{\mathrm{eff}}),
\]
where $k_0$ is the initial permeability, $k$ is the updated permeability, $\alpha$ is the stress sensitivity coefficient, and $\Delta \sigma_{\mathrm{eff}}$ is the change in effective stress.

Distinct values of $\alpha$ are assigned to different regions (matrix, propped fractures, and unpropped fractures) to account for their differing mechanical responses under depletion.

\subsubsection{Reservoir and Fracture Representation}

The reservoir is modeled as a three-dimensional system containing a hydraulically fractured horizontal well. Hydraulic fractures are represented as high-permeability planar features embedded within the reservoir grid, with local grid refinement applied near the wellbore and fracture network to resolve strong pressure and stress gradients.

The domain is partitioned into:
\begin{itemize}
    \item matrix regions,
    \item propped hydraulic fractures,
    \item unpropped fracture regions,
\end{itemize}
each characterized by distinct permeability and stress sensitivity properties. This representation enables accurate modeling of dominant flow pathways and their evolution during production.

\subsubsection{Governing Equations}

Fluid flow is governed by mass conservation combined with Darcy's law:
\[
\frac{\partial}{\partial t}(\phi \rho) + \nabla \cdot (\rho \mathbf{v}) = q,
\qquad
\mathbf{v} = -\frac{k}{\mu}\nabla p,
\]
where $\phi$ is porosity, $\rho$ is fluid density, $\mathbf{v}$ is Darcy velocity, $k$ is permeability, $\mu$ is viscosity, and $p$ is pressure.

Geomechanical behavior is modeled using linear elasticity, with stress equilibrium governed by:
\[
\nabla \cdot \boldsymbol{\sigma} + \mathbf{f} = 0,
\]
where $\boldsymbol{\sigma}$ is the stress tensor and $\mathbf{f}$ represents body forces. The constitutive relationship between stress and strain is defined using standard linear elastic parameters, including Young's modulus and Poisson's ratio.

\subsubsection{Boundary Conditions and Assumptions}

The model assumes linear elastic rock behavior and incorporates in-situ stress conditions representative of the reservoir. Boundary conditions include:
\begin{itemize}
    \item fixed displacement at the base of the domain,
    \item prescribed stresses at lateral and upper boundaries to approximate far-field confinement.
\end{itemize}

Flow and mechanical grids are aligned to ensure consistent pressure--stress coupling throughout the simulation.

\subsubsection{Stress Sensitivity Scenarios}

To evaluate the robustness of the optimization framework, multiple stress sensitivity scenarios are considered, corresponding to low, medium, and high values of the stress sensitivity coefficient $\alpha$. These scenarios represent different reservoir behaviors, ranging from weak to strong permeability degradation under depletion.

\subsubsection{Model Validation and Reuse}

\begin{figure}[htbp]
\centering
\includegraphics[width=\textwidth]{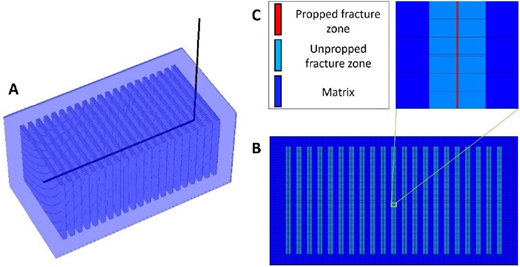}
\caption{The unconventional oil reservoir model with multistage hydraulic fractures: (a) 3D view of the model, (b) top view of the model, and (c) zoomed view of one hydraulic fracturing stage.}
\label{fig:reservoir_model}
\end{figure}

The model setup, calibration, and validation are described in detail in our previous work (Valiyev et al., 2025) and are adopted here without modification. A schematic representation of the reservoir model is shown in Figure~\ref{fig:reservoir_model}.

\subsection{Sampling Strategy}

Because the reliability of surrogate-based optimization depends strongly on the relationship between the training distribution and the trajectories encountered during optimization, the generation of training samples is a central component of the proposed framework.

The sampling strategy evolved through multiple stages. Initial datasets were generated using linear-decline-with-noise and moving-uniform sampling schemes. Intermediate datasets introduced variable-decline-with-noise trajectories. The final problem-informed strategy consisted of linear-decline, constant-or-decline, and combined trajectory classes designed to better match the structure of trajectories encountered during optimization.

\begin{figure}[htbp]
\centering
\includegraphics[width=\textwidth]{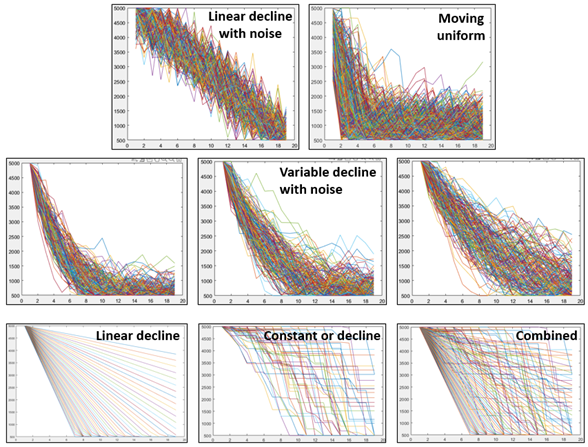}
\caption{Evolution of sampling strategies used for proxy training: initial linear-decline-with-noise and moving-uniform sampling schemes (top row), intermediate variable-decline-with-noise schemes (middle row), and the final problem-informed sampling strategy consisting of linear-decline, constant-or-decline, and combined trajectory classes (bottom row).}
\label{fig:sampling_strategy}
\end{figure}

This progression is illustrated in Figure~\ref{fig:sampling_strategy}. The final strategy was selected because it produces smoother, physically realistic, and optimization-relevant trajectories, thereby reducing extrapolation and improving surrogate robustness.

\subsection{Proxy Model Development}

To enable efficient optimization, a data-driven surrogate model is developed to approximate the mapping between time-varying bottomhole pressure (BHP) trajectories and cumulative oil production:
\[
J(\mathbf{u}) \approx \hat{J}(\mathbf{u}) = \hat{F}(\mathbf{u}),
\]
where $\mathbf{u} \in \mathbb{R}^T$ is the discretized control trajectory and $\hat{F}$ is a neural network surrogate.

\begin{figure}[htbp]
\centering
\includegraphics[width=0.85\textwidth]{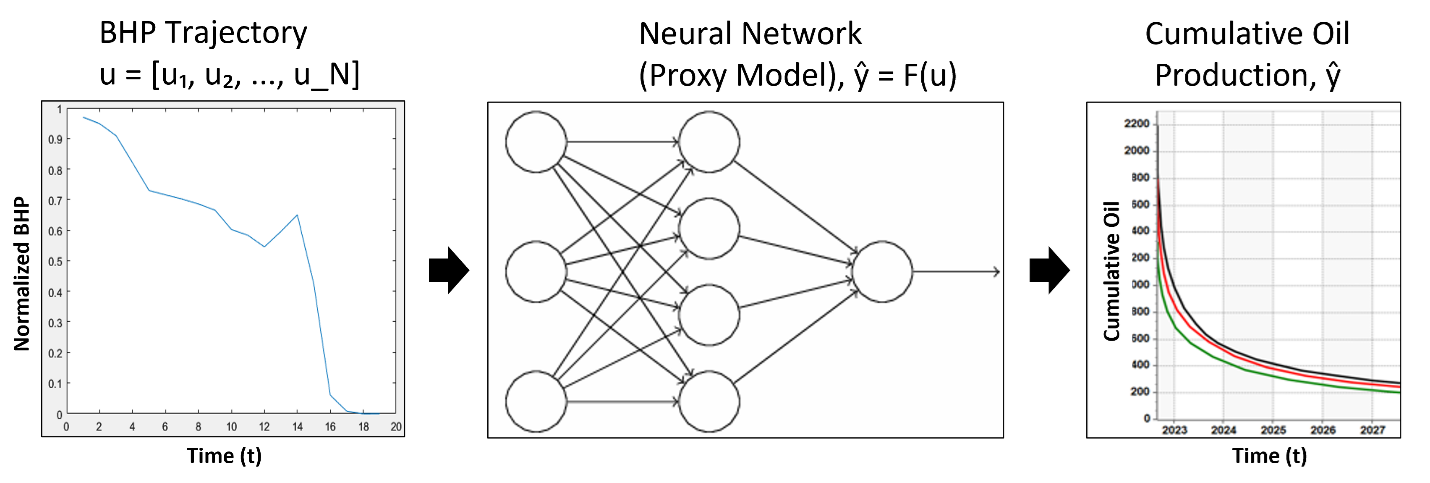}
\caption{Schematic representation of the proxy model mapping time-varying bottomhole pressure (BHP) trajectories to cumulative oil production. The neural network surrogate approximates the nonlinear relationship between control inputs and production response.}
\label{fig:proxy_model}
\end{figure}

A schematic representation of this mapping is shown in Figure~\ref{fig:proxy_model}, where the neural network approximates the nonlinear relationship between control inputs and production response. The figure illustrates the role of the surrogate model as a computationally efficient replacement for the full-physics simulator within the optimization workflow.

\subsubsection{Model Architecture}

A fully connected neural network (FCN) is adopted as the primary architecture. This choice is motivated by the fixed-dimensional structure of the control input and the need for computational efficiency during repeated evaluations within the optimization loop.

The input to the network is the normalized control trajectory:
\[
\tilde{\mathbf{u}} \in \mathbb{R}^T,
\]
and the output is the normalized cumulative oil production:
\[
\tilde{J}.
\]

In this study, $T \approx 20$, corresponding to a non-uniform temporal discretization with finer resolution at early production times.

The network consists of multiple fully connected layers with nonlinear activation functions, enabling approximation of the high-dimensional, nonlinear mapping between control trajectories and production response. A typical architecture consists of 3--4 hidden layers with moderate width, balancing representational capacity and generalization.

\subsubsection{Training Procedure}

The model is trained using supervised learning on the dataset
\[
\mathcal{D} = \left\{ \left(\mathbf{u}^{(i)}, J^{(i)} \right) \right\}_{i=1}^{N},
\]
where each sample is generated from the coupled flow--geomechanics simulator.

The loss function is defined as the mean squared error (MSE):
\[
\mathcal{L} = \frac{1}{N}\sum_{i=1}^{N} \left(\hat{J}^{(i)} - J^{(i)}\right)^2.
\]

Model performance is evaluated using standard regression metrics, including mean absolute error (MAE), root mean square error (RMSE), and coefficient of determination ($R^2$), computed on a held-out validation set.

\subsubsection{Data Efficiency and Generalization}

A key challenge in this problem is the limited size of the training dataset, as each sample requires a computationally expensive simulation. In this regime (typically 50--150 samples), predictive performance is governed more by the distribution and physical relevance of training data than by model complexity.

As discussed in the sampling strategy section and illustrated in Figure~\ref{fig:sampling_strategy}, structured and distribution-aligned datasets significantly improve surrogate accuracy compared to naive sampling approaches. The proxy model achieves highest accuracy on smooth, physically realistic trajectories that lie within the support of the training distribution.

This behavior is further demonstrated in Figure~\ref{fig:proxy_model_performance}, where proxy predictions are compared against simulator outputs. The model exhibits strong agreement for in-distribution samples, with predictions closely following the simulator response. In contrast, larger deviations are observed for irregular or out-of-distribution trajectories, highlighting the importance of distribution-aware sampling.

\subsubsection{Model Choice and Alternatives}

Alternative architectures, including sequence-based models such as recurrent neural networks, were explored but did not provide consistent performance improvements. This suggests that the temporal dependencies in the control sequence are sufficiently captured through the fixed discretization and do not require explicit sequential modeling.

Given the limited dataset size, increasing model complexity beyond a moderate fully connected architecture did not yield significant gains and, in some cases, degraded generalization performance. This reinforces the conclusion that data quality and distribution alignment are more critical than architectural sophistication in this setting.

\subsubsection{Reproducibility and Implementation Details}

To ensure reproducibility, detailed information on network architecture, hyperparameters, and training configuration is provided in Appendix A and Appendix B. These include layer configurations, activation functions, learning rate, batch size, and regularization settings.

\subsection{Optimization Framework}

We consider two optimization workflows for solving the high-dimensional well control problem:
\begin{enumerate}
    \item direct physics-based optimization, and
    \item surrogate (proxy)-based optimization.
\end{enumerate}

\begin{figure}[htbp]
\centering
\includegraphics[width=0.9\textwidth]{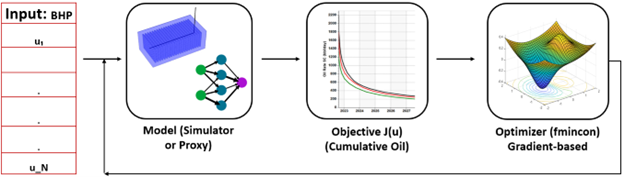}
\caption{Schematic representation of the optimization framework, showing iterative updates of control variables using a gradient-based optimizer (\texttt{fmincon}), where the objective function is evaluated using either the full-physics simulator or the neural network proxy model.}
\label{fig:optimization_framework}
\end{figure}

A schematic overview of both workflows is shown in Figure~\ref{fig:optimization_framework}, where the objective function is evaluated either using the full-physics simulator or the neural network surrogate within a gradient-based optimization loop.

\subsubsection{Physics-Based Optimization}

In the physics-based workflow, the objective function is evaluated directly using the coupled flow--geomechanics simulator. For a given control trajectory $\mathbf{u}$, cumulative oil production is obtained as
\[
J(\mathbf{u}) = Q_{\mathrm{oil}}^{\mathrm{cum}}(\mathbf{u}).
\]

Optimization is performed using a gradient-based constrained solver (MATLAB's \texttt{fmincon}), subject to the operational constraints defined in the problem formulation section.

Because the simulator does not provide analytical gradients, gradient information is approximated using finite-difference methods. As a result, each optimization iteration requires multiple simulator evaluations, leading to substantial computational cost. In practice, this results in hundreds of simulator runs per optimization, with total runtimes on the order of hours to days.

While this approach provides high-fidelity solutions, its computational expense limits its applicability for large-scale or repeated optimization tasks.

\subsubsection{Surrogate-Based Optimization}

To overcome this limitation, the simulator is replaced by the trained neural network surrogate:
\[
J(\mathbf{u}) \approx \hat{J}(\mathbf{u}) = \hat{F}(\mathbf{u}).
\]

The same constrained optimization problem is solved, but with the objective evaluated using the surrogate model. This enables rapid evaluation of candidate control trajectories, as each forward pass through the neural network is computationally inexpensive.

Unlike the simulator, the neural network surrogate is differentiable, allowing gradients of the objective function with respect to control variables to be computed efficiently via automatic differentiation. This significantly improves optimization efficiency and convergence behavior.

The same optimization algorithm (\texttt{fmincon}) is used in both workflows to ensure consistent handling of constraints, including:
\begin{itemize}
    \item bound constraints on bottomhole pressure,
    \item step-change constraints,
    \item optional monotonicity constraints.
\end{itemize}

This ensures that both physics-based and surrogate-based solutions are directly comparable in terms of feasibility and structure.

\subsubsection{Hybrid Validation Strategy}

To ensure the reliability of surrogate-based optimization, a hybrid validation step is incorporated into the workflow.

After the surrogate identifies an optimal or near-optimal control trajectory $\mathbf{u}^*$, the solution is re-evaluated using the full-physics simulator:
\[
J(\mathbf{u}^*) \quad \text{vs.} \quad \hat{J}(\mathbf{u}^*).
\]

The discrepancy between the surrogate prediction and the simulator result is used to assess the accuracy of the proxy in the optimized region. If the error exceeds a specified tolerance, additional simulator evaluations can be performed and incorporated into the training dataset, enabling refinement of the surrogate model.

This validation step mitigates the risk of optimization bias due to surrogate approximation errors, particularly in regions near the boundary of the training distribution.

\subsubsection{Proxy-Assisted Initialization}

In addition to serving as a standalone optimizer, the surrogate model can be used to generate high-quality initial guesses for physics-based optimization. In this hybrid approach:
\begin{enumerate}
    \item the surrogate model is used to rapidly identify a near-optimal control trajectory;
    \item the resulting trajectory is used as the initial condition for physics-based optimization.
\end{enumerate}

This significantly reduces the number of required simulator evaluations and improves convergence efficiency, combining the speed of surrogate-based optimization with the accuracy of full-physics simulation.

\subsubsection{Computational Considerations}

The primary advantage of the surrogate-based framework lies in its computational efficiency. While physics-based optimization requires repeated high-cost simulator evaluations, surrogate-based optimization replaces these with fast neural network evaluations.

As illustrated in Figure~\ref{fig:optimization_framework}, this substitution transforms the optimization workflow from a simulation-dominated process into a lightweight, computation-efficient procedure. This enables broader exploration of the control space and facilitates rapid evaluation of multiple scenarios.

\subsection{Validation and Robustness Analysis}

The reliability of the proposed surrogate-based optimization framework is evaluated through a series of validation and robustness tests designed to assess predictive accuracy, optimization consistency, and generalization capability across the control space.

The evaluation focuses on three key aspects:
\begin{enumerate}
    \item consistency of optimization outcomes across different initializations,
    \item agreement between surrogate predictions and full-physics simulator results,
    \item predictive performance on unseen control trajectories.
\end{enumerate}

\subsubsection{Sensitivity to Initialization}

To assess the stability of the optimization process, both physics-based and surrogate-based optimization workflows are initialized from multiple distinct bottomhole pressure (BHP) trajectories. These initial trajectories are selected to span different regions of the feasible control space, including both conservative and aggressive drawdown strategies.

For each initialization, optimization is performed independently, and the resulting control trajectories and cumulative production values are compared. This analysis evaluates whether the optimization framework consistently converges to similar high-quality solutions or exhibits sensitivity to initial conditions.

The resulting trajectories and convergence behavior are analyzed in the Results section (see Figures~\ref{fig:optimized_trajectories} and \ref{fig:problematic_cases}), where differences in optimization paths and final solutions are examined.

\subsubsection{Proxy Accuracy During Optimization}

To evaluate surrogate accuracy in the context most relevant to decision-making, proxy predictions are systematically compared against full-physics simulator outputs at two key stages:
\begin{itemize}
    \item \textbf{Initial evaluation:} For each initial control trajectory $\mathbf{u}_0$, the surrogate prediction $\hat{J}(\mathbf{u}_0)$ is compared with the simulator result $J(\mathbf{u}_0)$.
    \item \textbf{Final optimized solution:} For each optimized trajectory $\mathbf{u}^*$, the surrogate prediction $\hat{J}(\mathbf{u}^*)$ is compared with the simulator-evaluated outcome $J(\mathbf{u}^*)$.
\end{itemize}

This ensures that accuracy is assessed not only globally but also specifically in regions of the control space that influence optimization outcomes.

The comparison results are summarized quantitatively in Table~\ref{tab:proxy_validation_summary} and visualized through trajectory-level comparisons in Figure~\ref{fig:problematic_cases}.

\subsubsection{Generalization to Unseen Trajectories}

To evaluate generalization beyond the optimization path, the surrogate model is tested on control trajectories that are not used during training or optimization. These trajectories include both structured and perturbed profiles to probe different regions of the feasible control space.

Proxy predictions are compared against simulator outputs, and performance is analyzed in terms of both overall accuracy and sensitivity to trajectory characteristics. This evaluation is reflected in Figure~\ref{fig:proxy_model_performance}, where predicted and simulated production values are compared across representative cases.

\subsubsection{Error Metrics}

Prediction accuracy is quantified using relative error metrics defined as
\[
\text{Relative Error} = \frac{|\hat{J}(\mathbf{u}) - J(\mathbf{u})|}{J(\mathbf{u})}.
\]

These metrics are computed for both initial and optimized trajectories and are reported in Table~\ref{tab:proxy_validation_summary}. Additional performance measures, including MAE and RMSE, are used to assess global model accuracy (see the Proxy Model Development section).

Particular emphasis is placed on accuracy near optimal control trajectories, as these regions have the greatest impact on decision-making.

\subsubsection{Failure Mode Analysis}

In addition to aggregate accuracy metrics, the framework is evaluated through targeted analysis of cases exhibiting larger deviations between surrogate-based and physics-based optimization.

Representative failure cases are presented in Figure~\ref{fig:problematic_cases}, where discrepancies in both control trajectories and predicted production are observed. These cases are analyzed to identify underlying causes, including:
\begin{itemize}
    \item \textbf{Distribution mismatch:} optimization trajectories that lie near or outside the support of the training data distribution, leading to extrapolation;
    \item \textbf{Local optimization effects:} differences in optimization paths that result in convergence to alternative local optima.
\end{itemize}

These analyses provide insight into the limitations of the surrogate model and inform potential improvements through adaptive sampling or model refinement.

\section{Results and Discussion}

\subsection{Baseline Production Behavior and Impact of Stress Sensitivity}

Baseline simulations were performed to quantify the impact of stress sensitivity on production behavior prior to optimization. The results highlight the fundamental trade-off between pressure-driven flow and stress-induced permeability degradation that governs performance in unconventional reservoirs.

\begin{figure}[htbp]
\centering
\includegraphics[width=0.85\textwidth]{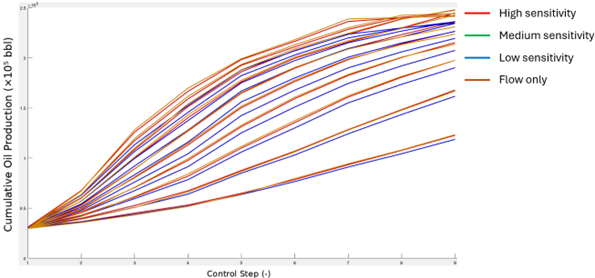}
\caption{Cumulative oil production as a function of control step for different stress sensitivity levels and flow-only conditions. Higher stress sensitivity leads to increased permeability degradation and reduced cumulative production, with differences becoming more pronounced at later control steps.}
\label{fig:baseline_production}
\end{figure}

As shown in Figure~\ref{fig:baseline_production}, cumulative oil production varies significantly across sensitivity levels. The flow-only case exhibits the highest production, as permeability remains constant and no geomechanical degradation occurs. In contrast, coupled simulations show reduced cumulative production due to stress-dependent permeability loss, with the magnitude of reduction increasing from low to high sensitivity cases.

A key observation is the temporal separation of these effects. At early control steps, production trends across sensitivity levels are similar, indicating that pressure-driven flow dominates. However, as production progresses, the impact of permeability degradation becomes increasingly pronounced, leading to divergence in cumulative production curves. This behavior demonstrates that aggressive drawdown strategies may appear favorable in short-term or flow-only analyses but lead to suboptimal outcomes when geomechanical effects are considered.

\subsection{Performance of Physics-Based Optimization}

Physics-based optimization was performed using the fully coupled simulator to determine optimal bottomhole pressure (BHP) trajectories under different stress sensitivity scenarios.

\begin{figure}[htbp]
\centering
\includegraphics[width=0.85\textwidth]{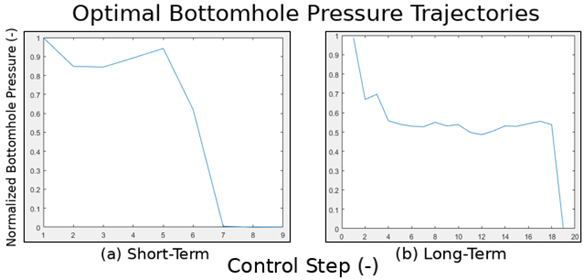}
\caption{Optimal bottomhole pressure (BHP) trajectories obtained from physics-based optimization for (a) short-term and (b) long-term production horizons. The results demonstrate non-linear, time-dependent control behavior driven by the trade-off between production enhancement and permeability preservation.}
\label{fig:optimized_trajectories}
\end{figure}

The optimized control trajectories, shown in Figure~\ref{fig:optimized_trajectories}, exhibit structured, nonlinear, and time-dependent behavior that cannot be captured by conventional predefined strategies. In short-term optimization scenarios (Figure~\ref{fig:optimized_trajectories}a), the optimal strategy favors higher initial drawdown to maximize early production, followed by rapid moderation to limit permeability degradation. In contrast, long-term optimization (Figure~\ref{fig:optimized_trajectories}b) results in more gradual drawdown profiles that maintain moderate pressure levels to preserve reservoir conductivity.

This behavior reflects the underlying trade-off between immediate production gains and long-term permeability preservation. The effect is more pronounced in medium and high stress-sensitivity cases, where permeability degradation has a stronger influence on production performance.

\subsection{Predictive Performance of the Proxy Model}

To enable efficient optimization, a neural network surrogate was developed to approximate the mapping between BHP trajectories and cumulative oil production.

\begin{figure}[htbp]
\centering
\includegraphics[width=0.85\textwidth]{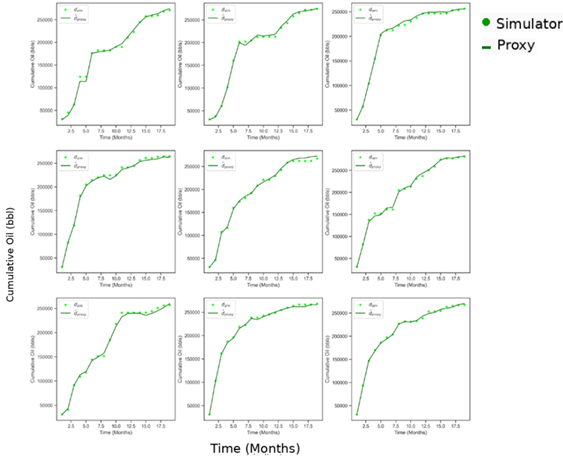}
\caption{Comparison between proxy-predicted and simulator-computed cumulative oil production. The proxy model captures the overall nonlinear relationship with good accuracy for in-distribution samples, while larger deviations occur for irregular or out-of-distribution trajectories.}
\label{fig:proxy_model_performance}
\end{figure}

As illustrated in Figure~\ref{fig:proxy_model_performance}, the proxy model captures the dominant nonlinear relationship between control inputs and production response with high fidelity for in-distribution samples. Predicted values closely follow simulator outputs across representative trajectories, and scatter plots (predicted vs.\ true values) exhibit strong alignment along the 45$^\circ$ line.

Quantitatively, the model achieves high predictive accuracy on validation data, with low relative error for trajectories consistent with the training distribution. However, performance degrades for irregular or out-of-distribution control patterns, where deviations from simulator results increase.

\subsection{Effect of Sampling Strategy on Proxy Accuracy}

The impact of training data distribution on surrogate performance is evaluated across multiple sampling strategies.

\begin{figure}[htbp]
\centering
\includegraphics[width=0.9\textwidth]{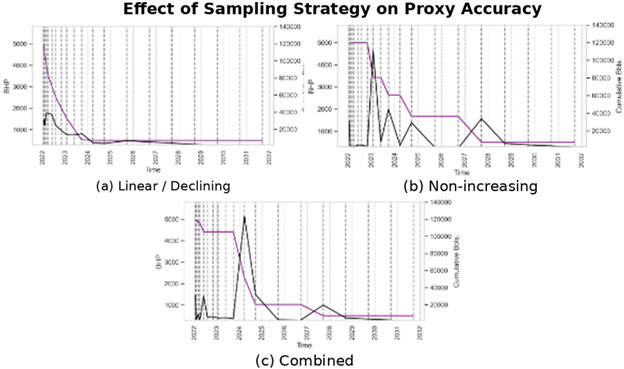}
\caption{Effect of training data sampling strategy on proxy model performance. (a) Linear/declining sampling captures general trends but lacks diversity. (b) Non-increasing sampling leads to larger deviations due to limited coverage. (c) Combined sampling provides the best agreement with simulator results.}
\label{fig:sampling_effect}
\end{figure}

As shown in Figure~\ref{fig:sampling_effect}, naive sampling approaches provide limited coverage of the control space and fail to capture the diversity of trajectories encountered during optimization. In contrast, the combined sampling strategy provides significantly improved performance by generating physically realistic and diverse trajectories aligned with the optimization search space.

These results are summarized quantitatively in Table~\ref{tab:sampling_strategy_comparison}, which shows that distribution-aligned sampling leads to lower prediction error and more stable optimization behavior.

\begin{table}[htbp]
\centering
\caption{Effect of sampling strategy on surrogate prediction error and distribution coverage.}
\label{tab:sampling_strategy_comparison}
\begin{tabular}{p{3.2cm} p{3.4cm} p{3cm} p{4.2cm}}
\toprule
Sampling Strategy & Distribution Coverage & Relative Error (Initial) & Relative Error (Final) \\
\midrule
Linear + noise & Partially out-of-distribution & $\sim$1--3\% & $>$5\% \\
Non-increasing & Frequently out-of-distribution & $\sim$2--5\% & $>$5\% \\
Piecewise & Mostly in-distribution & $<2\%$ & $<5\%$ \\
Combined & Fully in-distribution & $<2\%$ & $<5\%$ \\
\bottomrule
\end{tabular}
\end{table}

\subsection{Computational Efficiency: Proxy vs.\ Physics-Based Optimization}

A key advantage of the proposed framework is the substantial reduction in computational cost achieved through surrogate-based optimization.

As summarized in Table~\ref{tab:computational_cost_comparison}, physics-based optimization requires hundreds of expensive simulator evaluations, resulting in runtimes on the order of hours to days. In contrast, surrogate-based optimization replaces these evaluations with neural network predictions, reducing total optimization time to minutes.

\begin{table}[htbp]
\centering
\caption{Computational performance comparison between physics-based and proxy-based optimization.}
\label{tab:computational_cost_comparison}
\begin{tabular}{p{4cm} p{4cm} p{4cm}}
\toprule
Metric & Physics-Based & Proxy-Based \\
\midrule
Single simulation time & 5--18 minutes & $\sim$1 second \\
Number of simulations & $\sim$500 & 0 \\
Total optimization time & 1--2 days & 1--2 minutes \\
Function evaluations & Hundreds & Thousands \\
\bottomrule
\end{tabular}
\end{table}

\subsection{Validation of Proxy-Based Optimization}

To assess the reliability of surrogate-based optimization, optimized control trajectories obtained from the proxy model were re-evaluated using the full-physics simulator.

As shown in Table~\ref{tab:proxy_validation_summary}, strong agreement is observed between proxy-predicted and simulator-evaluated cumulative production across multiple initialization scenarios. Relative errors remain low for both initial and optimized trajectories.

\begin{table}[htbp]
\centering
\caption{Summary of surrogate-based optimization performance and validation results.}
\label{tab:proxy_validation_summary}
\begin{tabular}{p{4.5cm} p{7.5cm}}
\toprule
Metric & Observation \\
\midrule
Initial prediction error & $<2\%$ \\
Final solution error & $<5\%$ \\
Realized (simulated) error & $<5\%$ \\
Problematic cases & X1, X2, X6 \\
Trajectory agreement & Strong \\
Computational speedup & $\sim$1000$\times$ \\
Key limitation & Errors increase near training boundaries \\
\bottomrule
\end{tabular}
\end{table}

\subsection{Analysis of Problematic Cases}

While overall performance is robust, a small number of cases—specifically X1, X2, and X6—exhibit larger deviations between surrogate-based and physics-based optimization results.

\begin{figure}[htbp]
\centering
\includegraphics[width=0.9\textwidth]{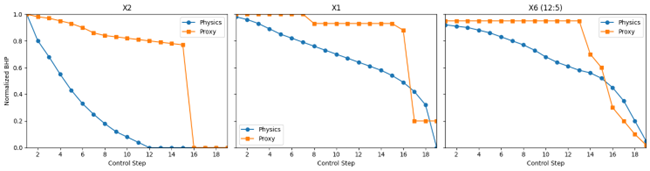}
\caption{Representative problematic cases in proxy-based optimization. (a) X1, (b) X2, and (c) X6 show increased deviations between proxy-based and physics-based trajectories due to boundary extrapolation and local optimization effects.}
\label{fig:problematic_cases}
\end{figure}

As illustrated in Figure~\ref{fig:problematic_cases}, these discrepancies are primarily associated with trajectories that lie near or outside the boundaries of the training data distribution. In such cases, the surrogate model operates in an extrapolation regime, leading to increased prediction error.

A secondary factor is local optimization behavior. In cases such as X2, the surrogate-based optimizer follows a different search path and converges to an alternative local optimum.

Despite these deviations:
\begin{itemize}
\item Overall trends and magnitudes remain consistent
\item Solutions remain near the optimal region
\item Proxy results provide strong initialization for refinement
\end{itemize}

\section{Conclusion}

In this study, we developed a deep learning--accelerated surrogate optimization framework for time-varying well control in stress-sensitive unconventional reservoirs. The proposed approach integrates a fully coupled flow--geomechanics model, a problem-informed sampling strategy, a neural network surrogate, and a constrained optimization workflow to enable efficient and reliable control optimization.

The results demonstrate that the surrogate model accurately approximates the nonlinear relationship between control trajectories and cumulative oil production in regions of the control space relevant to optimization. When embedded within the optimization loop, the surrogate-based approach achieves substantial computational speedup---on the order of several magnitudes---while maintaining close agreement with full-physics simulation results.

A key contribution of this work is the introduction of a distribution-aligned sampling strategy, which ensures that training data are representative of trajectories encountered during optimization. This significantly improves surrogate reliability and reduces extrapolation-driven error, enabling accurate optimization with a limited number of high-fidelity simulations.

The optimization results further show that optimal control strategies are inherently time-dependent and non-monotonic, reflecting the underlying trade-off between short-term production enhancement and long-term permeability preservation. These behaviors cannot be captured by predefined drawdown strategies, highlighting the importance of continuous, high-dimensional control formulations.

While the methodology is demonstrated in the context of coupled flow--geomechanics systems, the proposed framework is general and applicable to a broader class of high-dimensional optimization problems governed by computationally expensive simulators. By combining data-driven surrogates with physics-informed sampling and validation, the approach provides a scalable pathway for solving complex engineering optimization problems.

\section{Future Work}

Several directions can be pursued to further extend and enhance the proposed framework.

First, the sampling strategy can be improved through adaptive or active learning approaches, where new training samples are generated iteratively in regions of high prediction uncertainty or optimization relevance. This would further improve surrogate accuracy while reducing the number of required simulations.

Second, the current formulation focuses on cumulative oil production as the optimization objective. Future work can incorporate economic objectives, such as net present value (NPV), as well as multi-objective formulations that balance production, operational cost, and reservoir sustainability.

Third, the framework can be extended to more complex reservoir settings, including heterogeneous formations, naturally fractured systems, and uncertainty in reservoir properties. Incorporating uncertainty quantification would enable robust optimization under geological and operational uncertainty.

Fourth, alternative machine learning architectures, including sequence models and physics-informed neural networks, can be explored to better capture temporal structure and embed physical constraints directly into the surrogate model.

Finally, integration with reinforcement learning or real-time optimization strategies could enable adaptive control policies that evolve in response to changing reservoir conditions, further enhancing the practical applicability of the approach.

\section{References}

\begin{enumerate}

\item Almasoodi, M., Vaidya, R., and Reza, Z. (2020). Drawdown-management and fracture-spacing optimization in unconventional reservoirs. \textit{SPE Reservoir Evaluation \& Engineering}.

\item Boualam, A., Rasouli, V., Dalkhaa, C., and Djezzar, S. (2020). Stress-dependent permeability and porosity in Three Forks carbonate reservoir, Williston Basin. \textit{ARMA US Rock Mechanics/Geomechanics Symposium}.

\item Kumar, A., Seth, P., Shrivastava, K., and Sharma, M. M. (2018). Optimizing drawdown strategies in wells producing from complex fracture networks. \textit{SPE Hydraulic Fracturing Technology Conference}.

\item Liu, H. H., Rutqvist, J., and Berryman, J. G. (2009). On the relationship between stress and elastic strain for porous and fractured rock. \textit{International Journal of Rock Mechanics and Mining Sciences}, 46(2), 289--296.

\item Liu, H. H., Wei, M. Y., and Rutqvist, J. (2013). Normal-stress dependence of fracture hydraulic properties. \textit{Hydrogeology Journal}, 21(2), 371--382.

\item Mirani, A., Marongiu-Porcu, M., Wang, H., and Enkababian, P. (2018). Production-pressure-drawdown management for fractured horizontal wells. \textit{SPE Reservoir Evaluation \& Engineering}, 21(03), 550--565.

\item Morrill, J. C., and Miskimins, J. L. (2012). Optimizing hydraulic fracture spacing in unconventional shales. \textit{SPE Hydraulic Fracturing Technology Conference}.

\item Patzek, T. W., Male, F., and Marder, M. (2013). Gas production in the Barnett Shale obeys a scaling theory. \textit{PNAS}, 110(49), 19731--19736.

\item Quintero, J., and Devegowda, D. (2015). Modelling based recommendation for choke management in shale wells. \textit{SEG Global Meeting}.

\item Rojas, D., and Lerza, A. (2018). Horizontal well productivity enhancement in Vaca Muerta shale. \textit{SPE Canada Unconventional Resources Conference}.

\item Settari, A., Sullivan, R. B., and Bachman, R. C. (2002). Modeling water blockage and geomechanics. \textit{SPE Annual Technical Conference}.

\item Sun, Z., Liu, H. H., Han, Y., Basri, M. A., and Mesdour, R. (2021). Optimum pressure drawdown in shale gas reservoirs. \textit{Journal of Natural Gas Science and Engineering}, 88, 103848.

\item Tran, D., Nghiem, L., and Buchanan, L. (2009). Coupling reservoir flow and geomechanics. \textit{ARMA Symposium}.

\item Valiyev, M., Zheng, F., Liu, H. H., and Jafarpour, B. (2025). Optimization of well drawdown under geomechanical effects. \textit{URTEC}.

\item Wilson, K., and Hanna Alla, R. R. (2017). Efficient stress characterization for drawdown management. \textit{URTEC}.

\item Yu, J., and Jafarpour, B. (2022). Active learning for well control optimization. \textit{SPE Journal}, 27(05), 2668--2688.

\item Zhao, Y., et al. (2022). Choke management simulation for shale gas reservoirs. \textit{Journal of Natural Gas Science and Engineering}, 107, 104801.

\end{enumerate}

\appendix

\section{Neural Network Architecture and Training Details}

The neural network surrogate used in this study is a fully connected feedforward model designed to approximate the mapping between discretized bottomhole pressure (BHP) control trajectories and cumulative oil production.

\subsection{Model Configuration}

\begin{itemize}
    \item Input dimension: $T \approx 20$ (control steps)
    \item Output dimension: 1 (cumulative oil production)
    \item Architecture: Fully Connected Neural Network (FCN)
    \item Hidden layers: 3
    \item Neurons per layer: 64--128--64
    \item Activation function: ReLU
    \item Output activation: Linear
\end{itemize}

\subsection{Data Preprocessing}

Both input trajectories and output values are normalized using min--max scaling:
\[
\tilde{x} = \frac{x - x_{\min}}{x_{\max} - x_{\min}}.
\]

This ensures consistent scaling and improves numerical stability during training.

\subsection{Training Setup}

\begin{itemize}
    \item Loss function: Mean Squared Error (MSE)
    \item Optimizer: Adam
    \item Learning rate: $10^{-3}$
    \item Batch size: 8--16
    \item Epochs: 200--500 (with early stopping)
    \item Train/validation split: 80/20
    \item Regularization: Optional L2 weight decay
\end{itemize}

Early stopping is applied based on validation loss to prevent overfitting.

\subsection{Performance Characteristics}

The model achieves:
\begin{itemize}
    \item Low prediction error for in-distribution trajectories
    \item Stable training convergence
    \item Good generalization within the sampled control space
\end{itemize}

Performance degradation is observed for trajectories outside the training distribution, consistent with findings discussed in the main text (see Figures~\ref{fig:proxy_model_performance}--\ref{fig:problematic_cases}).

\section{Hyperparameter Tuning and Model Selection}

Hyperparameter tuning is performed to balance predictive accuracy and generalization, particularly under limited data conditions.

\subsection{Search Space}

The following hyperparameters are explored:

\begin{itemize}
    \item Number of hidden layers: 2--5
    \item Neurons per layer: 32--256
    \item Activation functions: ReLU, Tanh
    \item Learning rate: $10^{-4}$ -- $10^{-2}$
    \item Batch size: 8--32
    \item Regularization: None to small L2 weight decay
\end{itemize}

\subsection{Selection Procedure}

Each configuration is evaluated using a consistent train/validation split. Performance is assessed using:

\begin{itemize}
    \item Mean Absolute Error (MAE)
    \item Root Mean Square Error (RMSE)
    \item Coefficient of determination ($R^2$)
    \item Relative prediction error
\end{itemize}

The final model is selected based on the best trade-off between validation accuracy and generalization.

\subsection{Selected Configuration}

The final model corresponds to:

\begin{itemize}
    \item Hidden layers: 3
    \item Neurons: 64--128--64
    \item Activation: ReLU
    \item Optimizer: Adam
    \item Learning rate: $10^{-3}$
    \item Batch size: 8--16
    \item Epochs: $\sim$300 (with early stopping)
    \item Regularization: small L2 weight decay
\end{itemize}

\end{document}